\documentclass[conference]{IEEEtran}
\IEEEoverridecommandlockouts
\usepackage{amsmath, float, subcaption, graphicx, multirow, diagbox}

\graphicspath{ {figs/} }

\begin{document}
\title{Lightweight Image Codec via Multi-Grid Multi-Block-Size Vector 
Quantization (MGBVQ)}

\author{\IEEEauthorblockN{
Yifan Wang\IEEEauthorrefmark{1},
Zhanxuan Mei\IEEEauthorrefmark{1}, 
Ioannis Katsavounidis\IEEEauthorrefmark{2} and 
C.-C. Jay Kuo\IEEEauthorrefmark{1}}
\IEEEauthorblockA{\IEEEauthorrefmark{1}University of Southern California, Los Angeles, California, USA}
\IEEEauthorblockA{\IEEEauthorrefmark{2} Facebook, Inc., Menlo Park, California, USA}}

\maketitle

\begin{abstract}

A multi-grid multi-block-size vector quantization (MGBVQ) method is proposed for image coding in this work. The fundamental idea of image coding is to remove correlations among pixels before quantization and entropy coding, e.g., the discrete cosine transform (DCT) and intra predictions, adopted by modern image coding standards. We present a new method to remove pixel correlations. First, by decomposing correlations into long- and short-range correlations, we represent long-range correlations in coarser grids due to their smoothness, thus leading to a multi-grid (MG) coding architecture. Second, we show that short-range correlations can be effectively coded by a suite of vector quantizers (VQs). Along this line, we argue the effectiveness of VQs of very large block sizes and present a convenient way to implement them. It is shown by experimental results that MGBVQ offers excellent rate-distortion (RD) performance, which is comparable with existing image coders, at much lower complexity. Besides, it provides a progressive coded bitstream. 

\end{abstract}

\IEEEpeerreviewmaketitle

\section{Introduction}\label{sec:introduction}
Image coding has been studied for more than four decades. Many image coding standards have been finalized and widely used today, e.g., JPEG \cite{wallace1992jpeg} and intra coding of many video coding standards, e.g., H.264 intra \cite{wiegand2003overview}, Webp \cite{lian2012webp} from VP8 intra \cite{bankoski2011technical}, BPG \cite{bpg} from HEVC intra \cite{sullivan2012overview}, AV1 intra \cite{chen2018overview} and VVC intra \cite{wang2021high}. To achieve higher and higher coding gains, the coding standards become more and more complicated by including computationally intensive units like intra prediction, flexible coding block partition and more transform types.  Specialized hardware are developed to reduce the coding time  \cite{albalawi2015hardwarebpg, elbadri2005hardware}.

Another new development is the emergence of learning-based image coding solutions \cite{yang2015estimating, yang2020learned}.  Image coding methods based on deep learning \cite{goodfellow2016deep} have received a lot of attention in recent years, due to their impressive coding gain.  For example, Balle et. al \cite{balle2016end} used the variational autoencoder (VAE) \cite{kingma2019introduction}, generalized divisive normalization (GDN) \cite{balle2015density}, the hyper-prior model \cite{balle2018variational} and a carefully designed quantization scheme to train an end-to-end optimized image codec. Recent work on deep-learning-based (DL-based) image/video coding achieves nice results and were reviewed in \cite{liu2020deep,ma2019image}.

Recent image coding RnD activities have focused on rate-distortion (RD) \cite{sullivan1998rate, choi2006fast} performance improvement at the expense of higher complexity. Here, we pursue another direction by looking for a lightweight image codec even at the cost of slightly lower RD performance. In particular, we revisit a well-known learning-based image coding solution, i.e., vector quantization (VQ). There has been a long history of VQ research \cite{gray1984vector} and its application to image coding \cite{nasrabadi1988image,mohamed1995image,huang1990fast}. Yet, VQ-based image codec has never become mainstream. To make VQ a strong competitor, it is essential to cast it in a new framework. 

A multi-grid multi-block-size vector quantization (MGBVQ) method is proposed for image coding in this work. The fundamental idea of image coding is to remove correlations among pixels before quantization and entropy coding. For example, DCT and intra predictions are adopted by modern image coding standards to achieve this objective. To remove pixel correlations, our first idea is to decompose pixel correlations into long- and short-range correlations. Long-range correlations can be represented by coarser grids since they are smoother. Recursive application of this idea leads to a multi-grid (MG) coding architecture.  Our second idea is to encode short-range pixel correlations by a suite of vector quantizers (VQs). Along this line, we argue the effectiveness of VQs of very large block sizes and present a convenient way to implement them.  Experiments demonstrate that MGBVQ offers an excellent rate-distortion (RD) performance that is comparable with state-of-the-art image coders at much lower complexity. 

Furthermore, progressive coding \cite{tzou1987progressive,malvar2000fast} is a desired functionality of an image codec. For example, JPEG has a progressive mode.  JPEG-2000 \cite{rabbani2002jpeg2000} also supports progressive coding.  H.264/SVC offers a scalable counterpart of H.264/AVC \cite{stutz2011survey}, which is achieved by sacrificing the RD performance \cite{wien2007performance}.  We will show that MGBVQ offers a progressive coding bitstream by nature.  

The rest of this paper is organized as follows. The MGBVQ method is introduced in Sec.  \ref{sec:Methods}. Experimental results are shown in Sec.  \ref{sec:experiment}. Concluding remarks and future research directions are pointed out in Sec. \ref{sec:conclusion}. 

\begin{figure}[tbp]
\centering
\begin{subfigure}[b]{0.49\textwidth}
\centering
\includegraphics[width=0.81\textwidth]{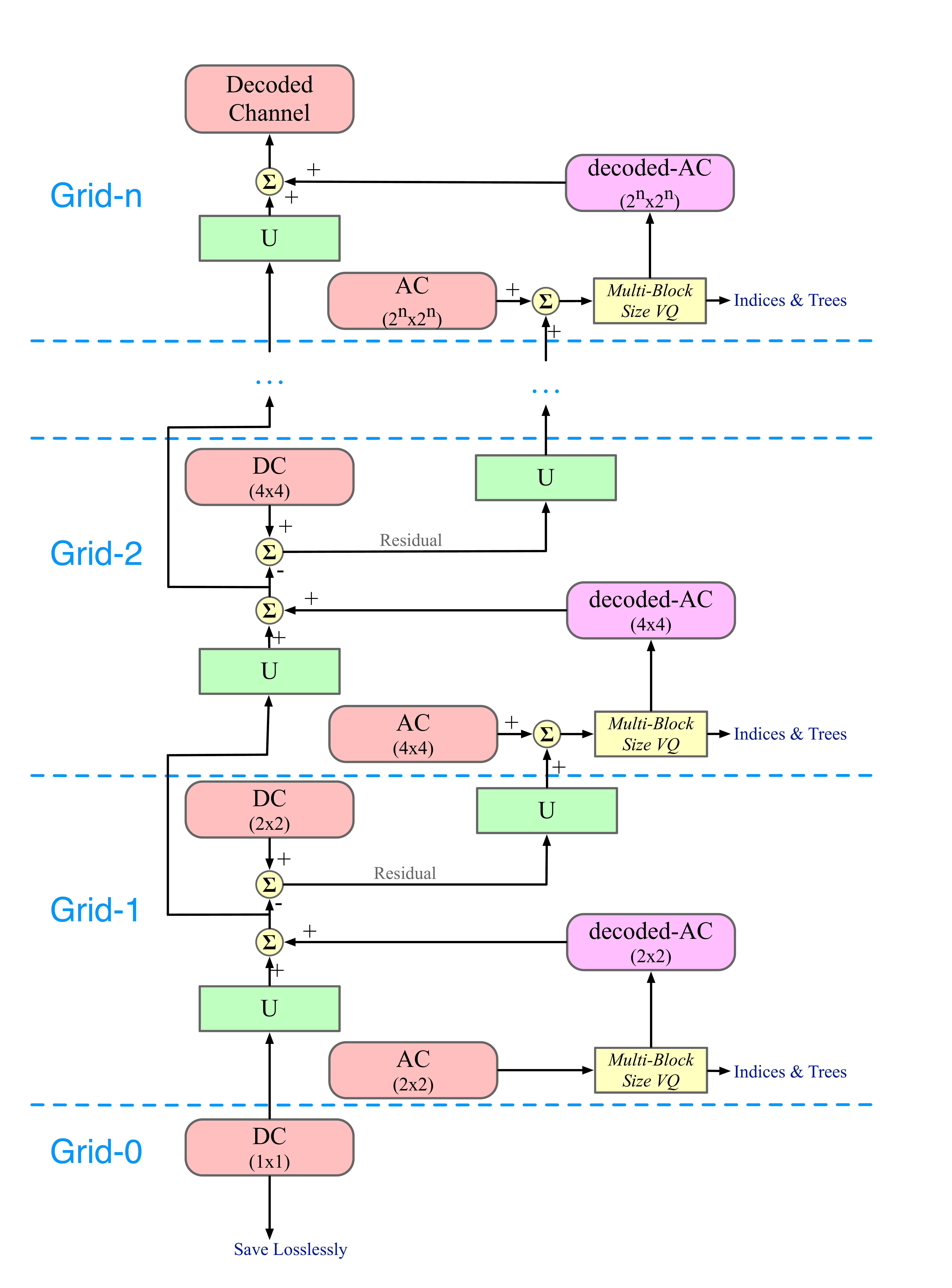}
\end{subfigure}
\caption{An overview of the proposed MGBVQ method.}\label{fig:backward}
\end{figure}

\section{MGBVQ Method}\label{sec:Methods}

\subsection{Multi-Grid (MG) Representation} 

To encode/decode images of spatial resolution $2^N\times2^N$, we build a
sequence of grids, $G_n$, of spatial resolution $2^n\times2^n$, where
$n=1, \cdots, N$, to represent them as shown in Fig. \ref{fig:backward}.
To achieve this goal, we begin with an image, $I_N$, at the finest grid
$G_N$. We decompose it into a smooth component and a residual component.
The latter is the difference between the original signal and the smooth
component. For convenience, the smooth and the residual components are
called DC and AC components for the rest of this paper. Since the DC
component is smooth, we downsample it using the Lanczos operator
\cite{lanczos1964evaluation} and represent it in grid $G_{N-1}$.
Mathematically, this can be written as
\begin{equation}\label{eq:downsampling} 
I_{N-1} = D(I_N),
\end{equation}
where $D$ denotes the downsampling operation from $G_N$ to $G_{N-1}$.
Similarly, we can upsample image $I_{N-1}$ to $I_N$ via 
\begin{equation}\label{eq:upsampling} 
\tilde{I}_{N} = U (I_{N-1}) = U [ D (I_N) ].
\end{equation}
where the upsampling operation, $U$, is also implemented by the Lanczos 
operator. Then, the AC component can be computed by
\begin{equation}\label{eq:AC} 
AC_{N} = I_N - \tilde{I}_{N} = I_N - U [ D (I_N) ].
\end{equation}
The above decomposition process can be repeated for $I_{N-1}$,
$I_{N-2}$, up to $I_2$.  Then, grid $G_n$ is used to represent $AC_n$,
where $n=N, N-1, \cdots, 2$ while grid $G_0$ is used to represent $DC_0$
in the MG image representation. 

It is worthwhile to comment that all existing image/video coding
standards adopt the single-grid representation.  The difference between
smooth and textured/edged regions is handled by block partitioning with different block sizes. Yet, there are low-frequency components in
textured/edged regions and partitioning these regions into smaller blocks has a negative impact on exploiting long-range pixel correlations, which can be overcome by the MG representation.  Another
advantage of the MG representation is its effectiveness in the RD
performance.  Suppose that grid $G_n$ pays the cost of one bit per pixel
(bpp) to reduce a certain amount of mean-squared error (MSE) denoted by
$\Delta MSE$. The cost becomes $4^{n-N}$ bpp at grid $G_N$ since the MSE
reduction is shared by $4^{N-n}$ pixels. 

\begin{table}\centering\scriptsize
\caption{Parameter setting for $256\times 256$ images, where we specify the number 
of codewords and the number of spectral components in codebook $C_{n,m}$ by 
(\#codeword, \#components). For $C_{8,3}$, we have $N\in \{8,16,32,64,128\}$
which gives the 5 points in Fig. \ref{fig:clic256}}\label{table:par256}
\begin{tabular}{ccccccc} \hline
        & $G_8$     & $G_7 $    & $G_6$     & $G_5$     & $G_4$     & $G_3$ \\ \hline
$C_{*,8}$     &(64,150)     & -         & -         & -         & -         & -     \\
$C_{*,7}$     &(128,150)    & (64,150)    & -         & -         & -         & -      \\
$C_{*,6}$      &(512,150)    & (128,150)    & (64,100)    & -         & -         & -      \\
$C_{*,5}$      & (512,50)    & (512,50)    & (128,40)    & (64,40)    & -         & -      \\
$C_{*,4}$      & (512,30)    & (512,30)    & (512,20)    & (128,20)    & (64,20)    & -      \\
$C_{*,3}$       & (N,-)    & (512,20)     & (512,12)    & (512,12)    & (32,12)    & (64,12) \\
$C_{*,2}$       & -         & (128,-)     & (512,-)    & (512,-)    & (64,-)    & (32,-) \\ \hline
\end{tabular}

\end{table}
   
\begin{figure*}[ht]
     \centering
     \begin{subfigure}[b]{0.32\textwidth}
        \centering
         \includegraphics[width=0.99\textwidth,height=0.65\textwidth]{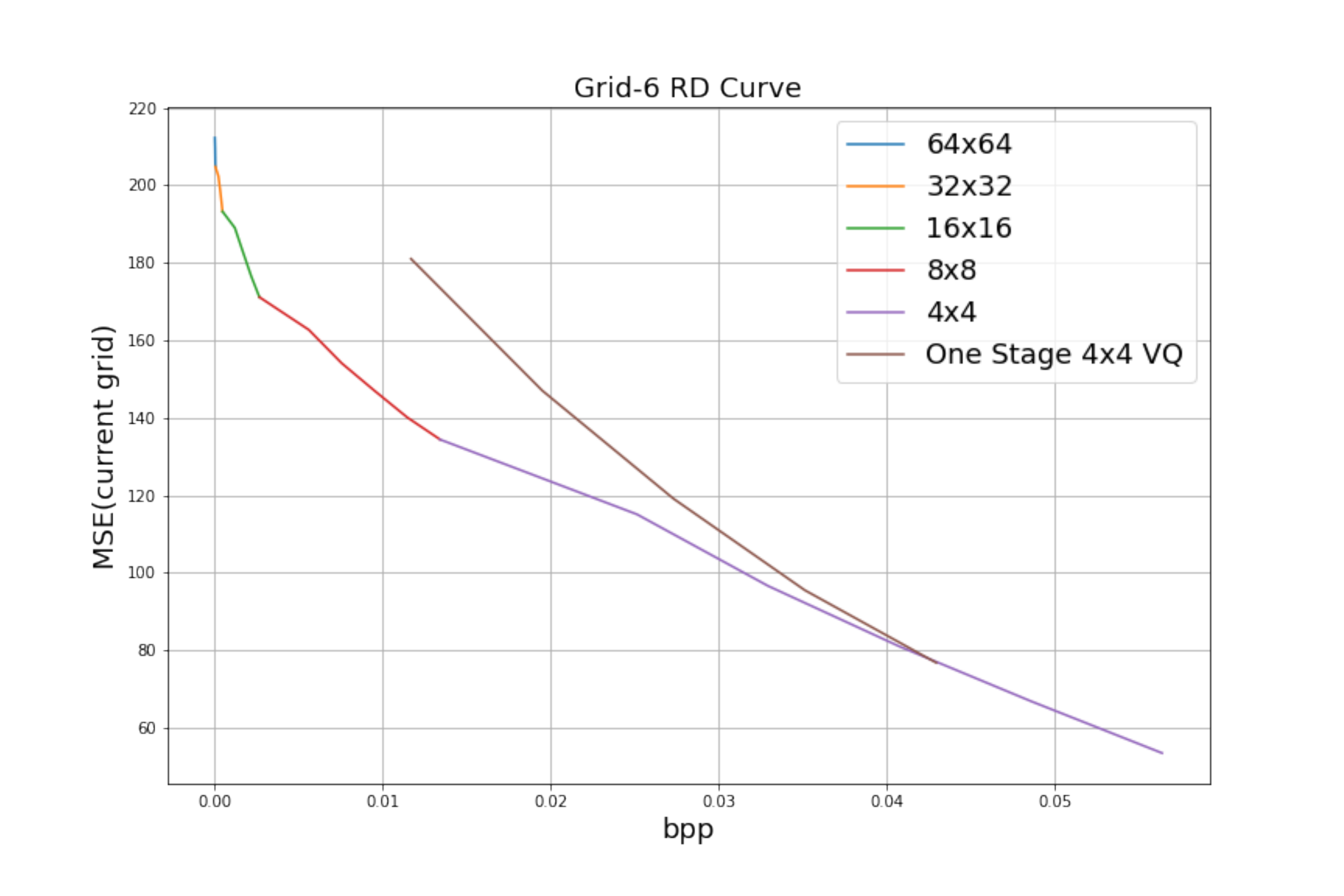}
         \caption{$G_6$ RD curve}\label{fig:rd1}
     \end{subfigure}
     \begin{subfigure}[b]{0.32\textwidth}
        \centering
         \includegraphics[width=0.99\textwidth,height=0.65\textwidth]{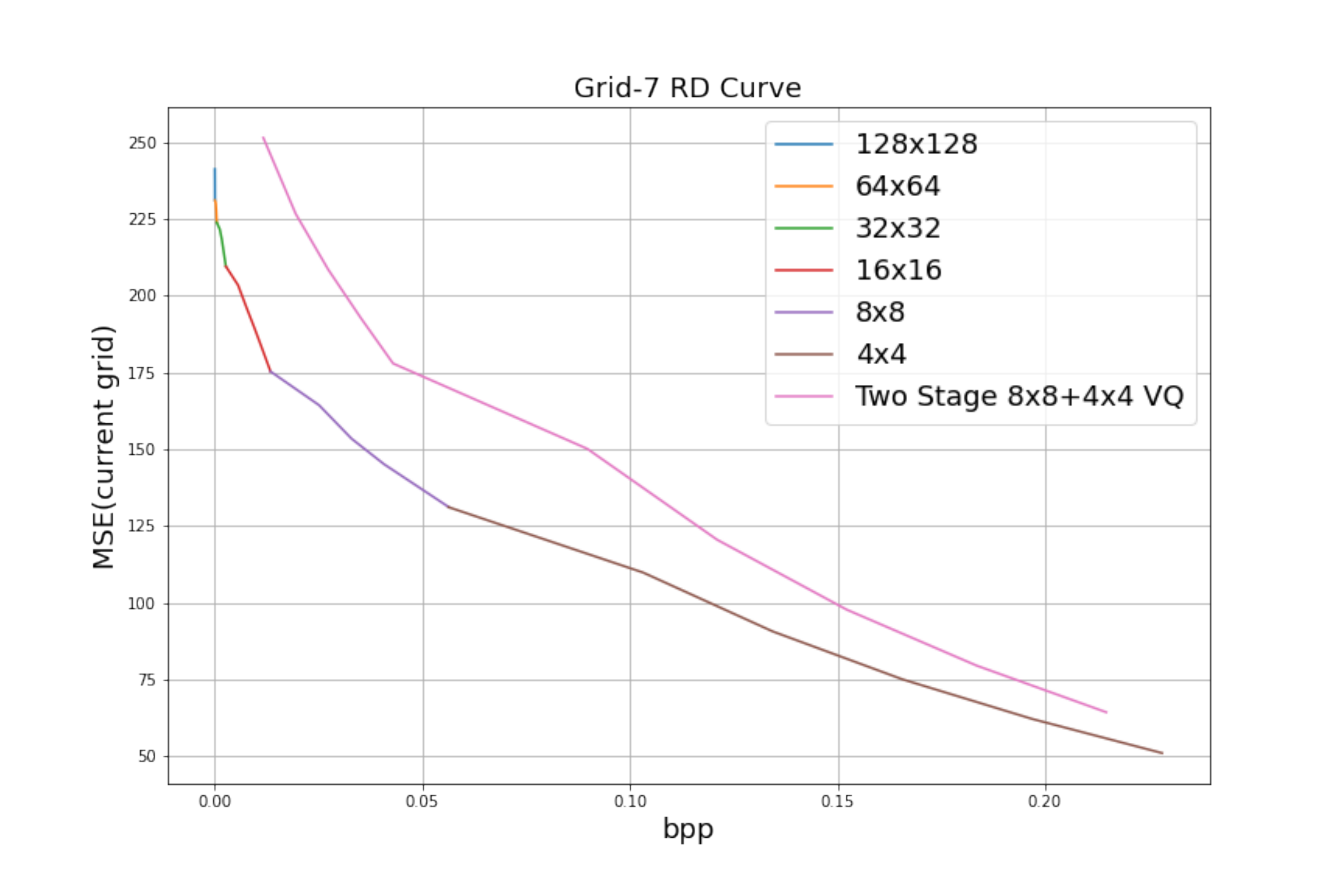}
         \caption{$G_7$ RD curve}\label{fig:rd2}
     \end{subfigure}
      \begin{subfigure}[b]{0.32\textwidth}
        \centering
         \includegraphics[width=0.99\textwidth,height=0.65\textwidth]{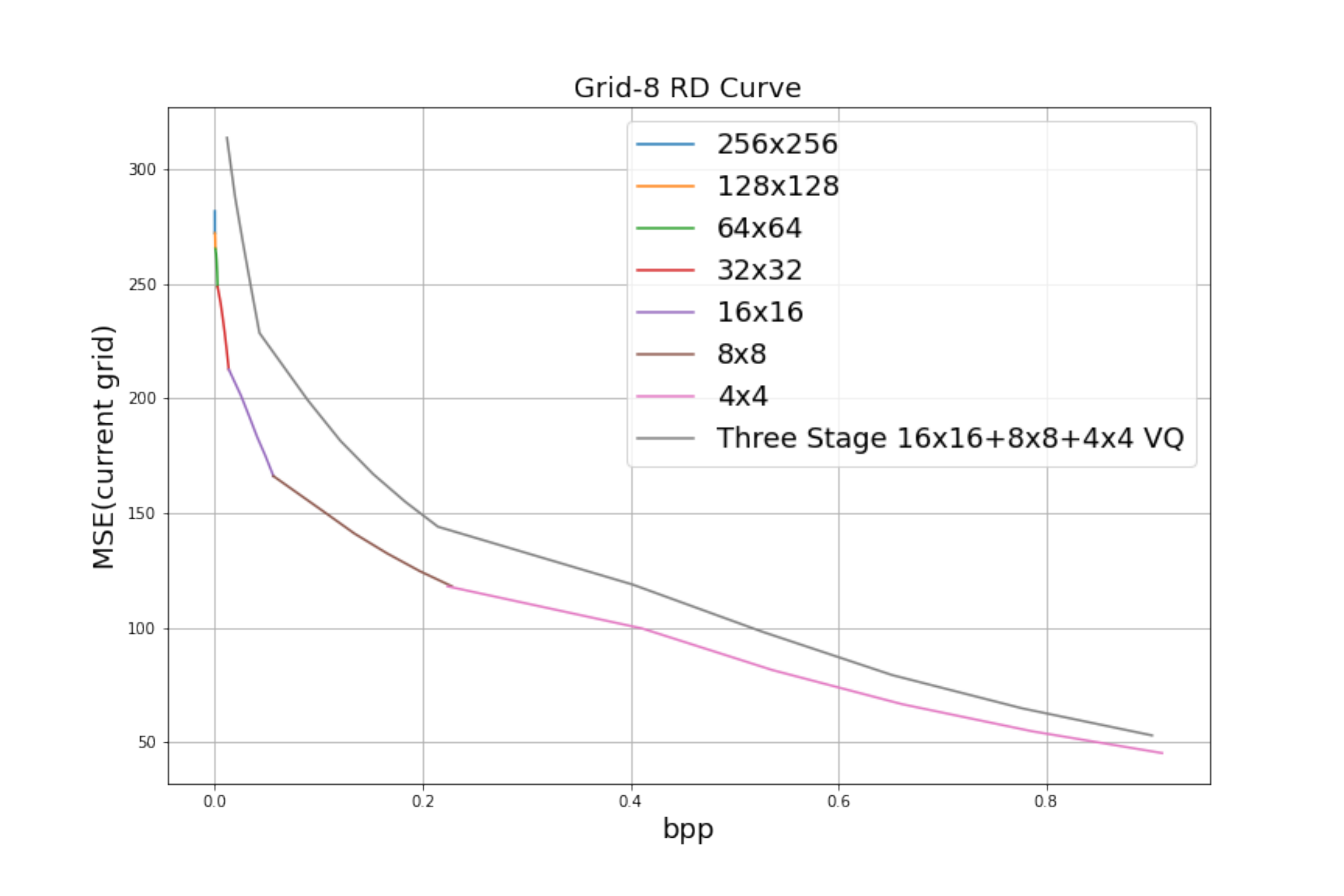}
         \caption{$G_8$ RD curve}\label{fig:rd3}
     \end{subfigure}
\caption{The RD curves for grids $G_6$, $G_7$ and $G_8$, which are benchmarked 
with one-stage, two-stage and three-stage VQ designs.}\label{fig:rd}
\end{figure*}

\subsection{Multi-Block-Size VQ}\label{subsec:mb2vq}

We use VQ to encode AC components in grids $G_n$, $n=2, \cdots, N$. Our
proposed VQ scheme has several salient features as compared with the
traditional VQ. That is, multiple codebooks of different block sizes (or
codeword sizes\footnote{We use codeword sizes and block sizes
interchangeable below.}) are trained at each grid.  The codebooks are
denoted by $C_{n,m}$, where subscripts $n$ and $m$ indicate the index of
grid $G_n$ and the codeword dimension ($2^m \times 2^m$), respectively.
We have $m \leq n$ since the codeword dimension is upper
bounded by the grid dimension. We use $|C_{n,m}|$ to denote the number
of codewords in $C_{n,m}$. 

At grid $G_n$, we begin with codebook $C_{n,n}$ of the largest codeword
size, where the codeword size is the same as the grid size. As a result,
the AC component can be coded by one codeword from the codebook $C_{n,n}$.
After that, we compute the residual between the original and the
$C_{n,n}$-quantized AC images, partition it into four non-overlapping
regions, and encode each region with a codeword from codebook
$C_{n,n-1}$.  This process is repeated until the codeword size is
sufficiently small with respect to the target grid. 

Before proceeding, it is essential to answer the following several questions: 
\begin{itemize}
\item Q1: How to justify this design?
\item Q2: How to train VQ codebooks of large codeword sizes? How to 
encode an input AC image with trained VQ codebooks at Grid $G_n$?
\item Q3: How to determine codebook size $|C_{n,m}|$?
\item Q4: When to switch from codebook $C_{n,m}$ to codebook $C_{n,m-1}$?
\end{itemize}

VQ learns the distribution of samples in a certain space, denoted by
$R^L$, and finds a set of representative ones as codewords. The error
between a sample and its representative one has to be normalized by
dimension $L$. If there is an upper bound of the average error (i.e.
the MSE), a larger $L$ value is favored. On the other hand, there is
also a lower bound on the average error since the code size cannot be too
large (due to worse RD performance). To keep good RD performance, we
should switch from codebook $C_{n,m}$ to codebook $C_{n,m-1}$. Actually,
Q1, Q3 and Q4 can be answered rigorously using the RD analysis. This
will be presented in an extended version of this paper. Some RD results
of codebooks of different codeword sizes are shown in the experimental
section. 

To answer Q2, we transform an image block of sizes $2^m \times 2^m$ from
the spatial domain to the spectral domain via a sequence of cascaded
channel-wise (c/w) Saab transforms \cite{Chen2020pixelhop++}.  The c/w
Saab transform is a variant of the principal component analysis (PCA).
However, a single-stage PCA is difficult to apply for large $m$.
Instead, we group small non-overlapping blocks (say, of size $2 \times
2$) and conduct the transform locally. Then, we discard channels of low
response values and keep channels of high response values for dimension
reduction. Finally, we are able to reach the final stage that has only a
spatial dimension of $1 \times 1$ of $K$ spectral components.  The
combination of multi-stage local filtering and dimension reduction
allows the spectral analysis of any blocks of large sizes. All transform
parameters are learned from the data. We train codebook $C_{n,n}$ based
on clustering of $K$ components in the spectral domain. After
quantization, we can transform the quantized codeword back to the
spatial domain via the inverse c/w Saab transform. This can be
implemented as a look-up table since it has to be done only once. Then,
we can compute the AC residual by subtracting the quantized AC value from 
the original AC value.

There are miscellaneous coding tools used.  First, for color image
coding, it is observed that the c/w Saab transform can compact energy
well in the spatial domain as well as R, G, B three channels. Thus, when
we train VQ codebooks $C_{n,m}$, we simply consider codewords of
dimension $2^m \times 2^m \times 3$.  Second, the simple Huffman coder
is adopted as the entropy coder. The codeword indices of each codebook
$C_{n,m}$ are coded using a Huffman table learned from training data.
Third, it is advantageous to adopt different codebooks in different
spatial regions at finer grids. For example, we can switch to codebooks
of smaller block sizes in complex regions while stay with codebooks of
larger blocks in simple regions. This is a rate control problem. We
implement rate control using a quad-tree. If a region is smooth, there
is no need to switch to codebooks of smaller code sizes.  This is called
early termination. Early termination helps reduce the number of bits a
lot in finer grids. For example, without early termination, the coding
of Lena of resolution $256\times 256$ at $G_8$ with $C_{8,8}$, which has
64 codewords, needs 0.09bpp. Early termination with entropy coding can
reduce it to 0.0246bpp. 

\begin{figure*}[t]
     \centering
     \begin{subfigure}[b]{0.22\textwidth}
        \centering
         \includegraphics[width=0.85\textwidth]{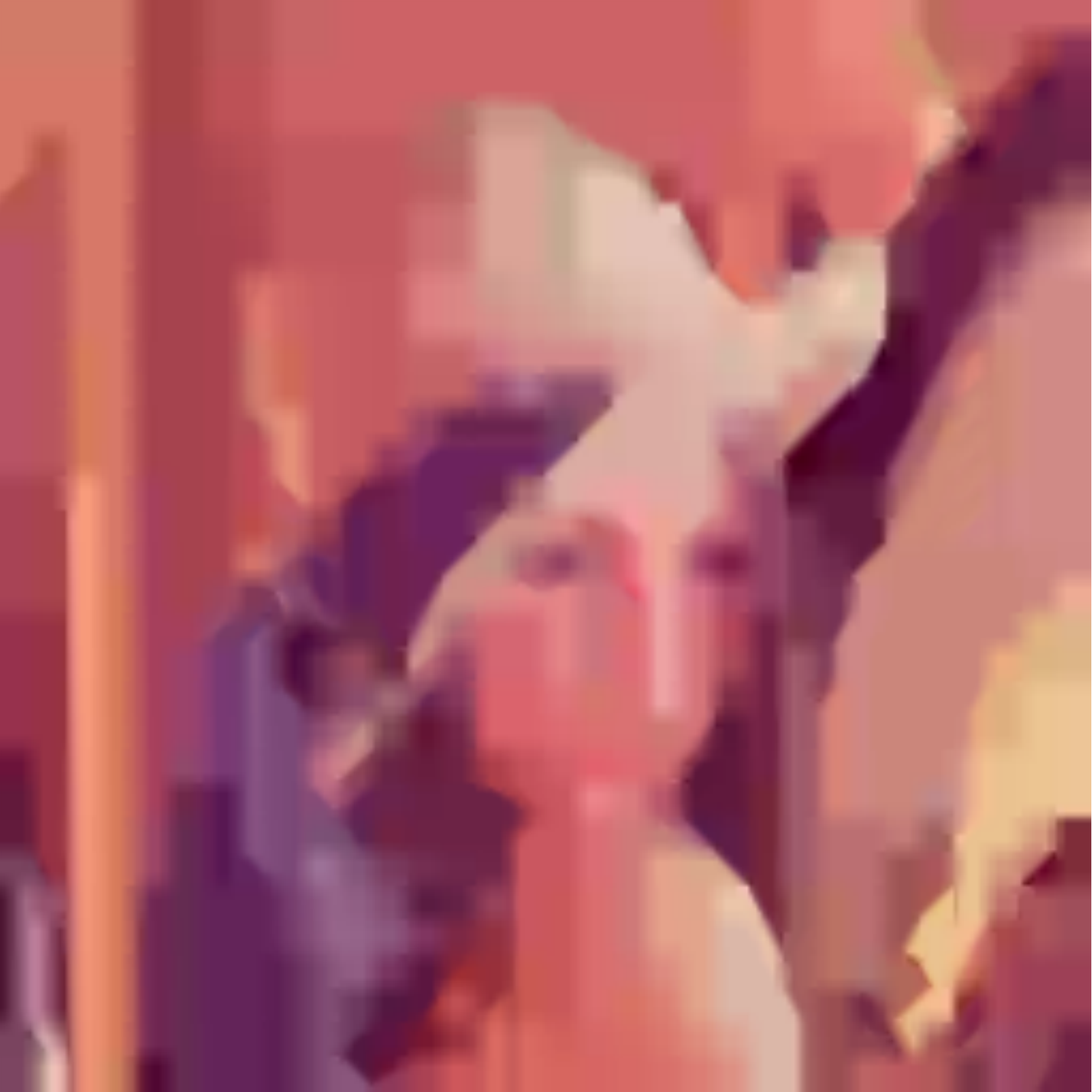}
         \caption{\centering H.264-444  $21.38dB@0.2447bpp$}\label{fig:lena_h264256}
     \end{subfigure}
      \begin{subfigure}[b]{0.22\textwidth}
        \centering
         \includegraphics[width=0.85\textwidth]{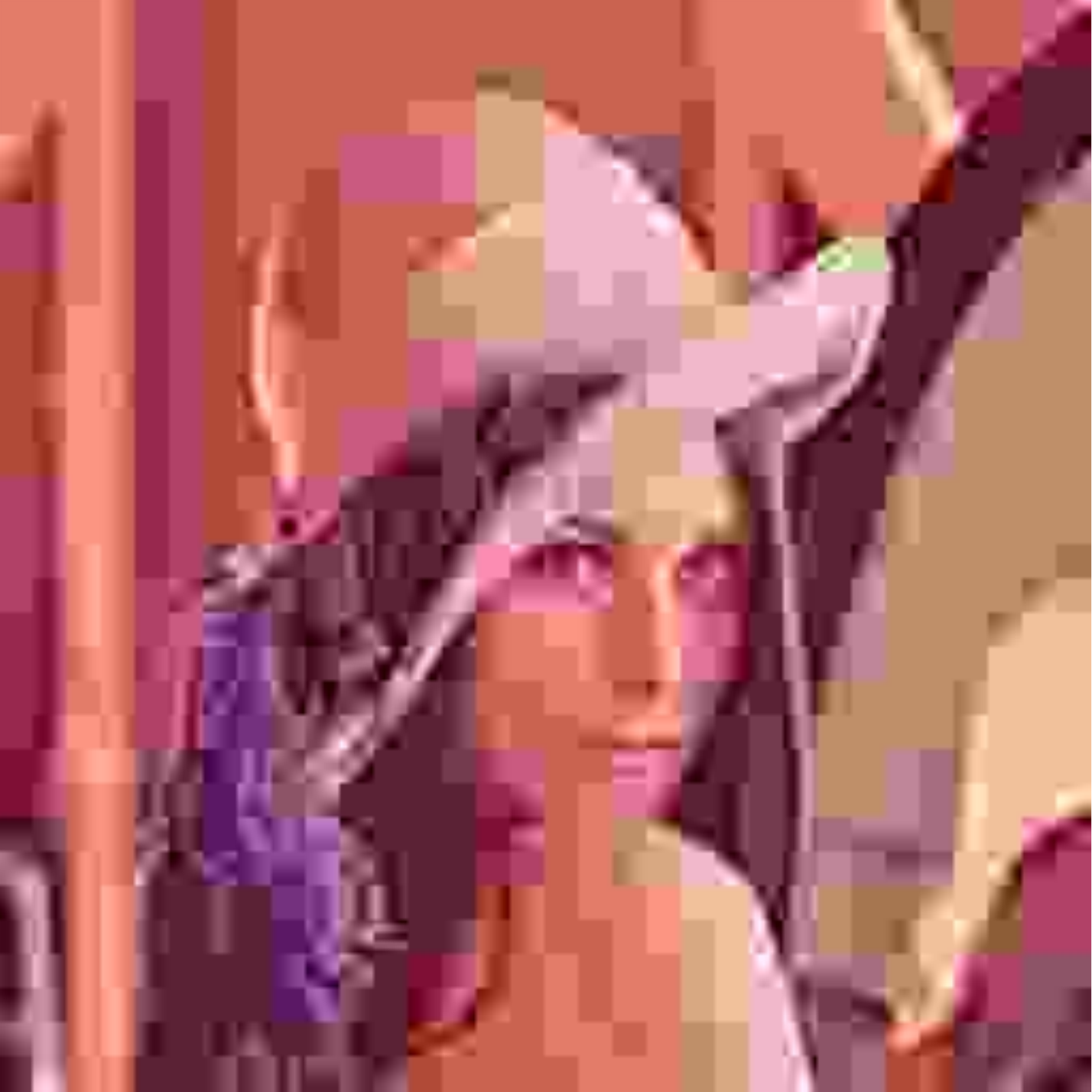}
         \caption{\centering JPEG Progressive  $22.21dB@0.2355bpp$}\label{fig:lena_jpg_prog256}
     \end{subfigure}
     \begin{subfigure}[b]{0.22\textwidth}
        \centering
         \includegraphics[width=0.85\textwidth]{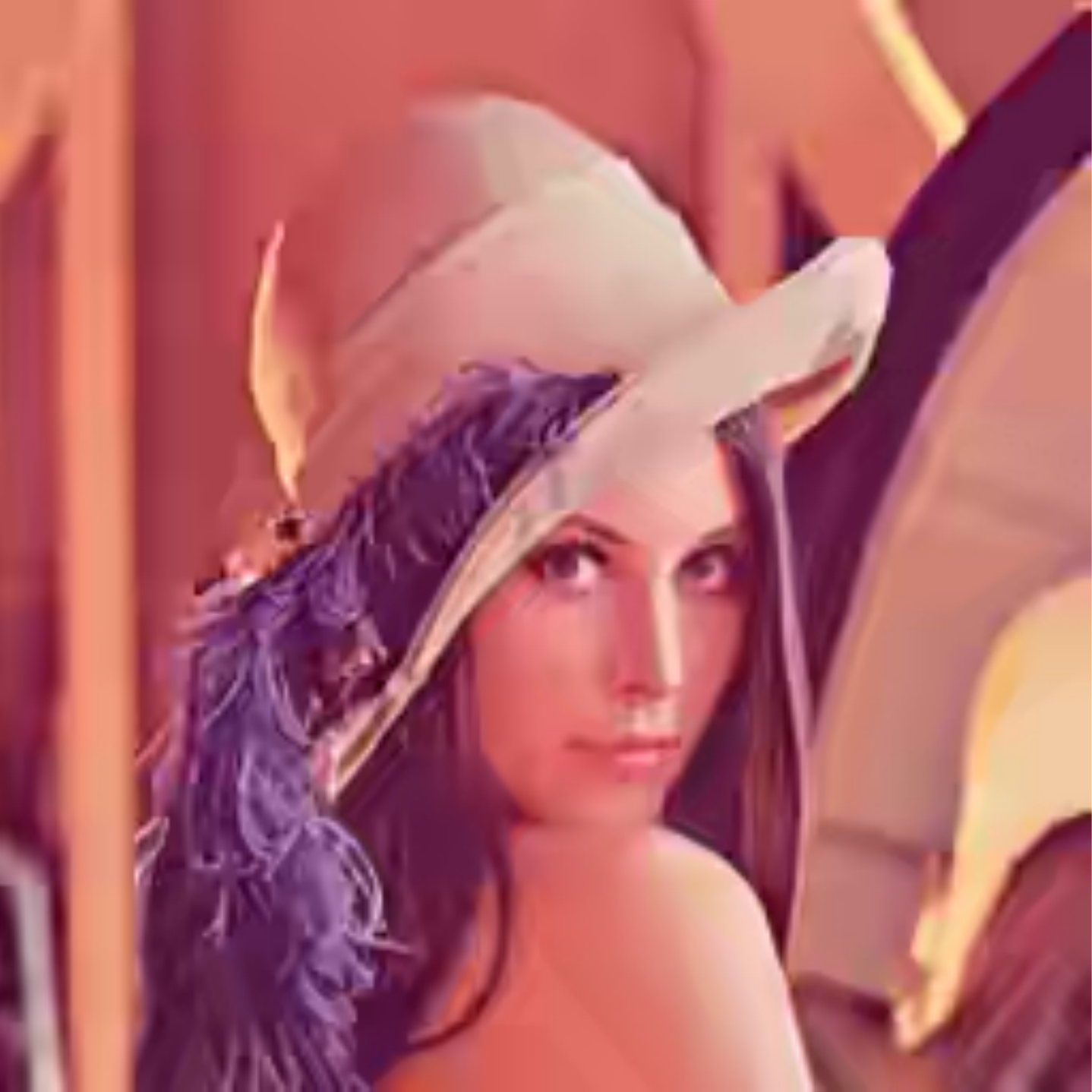}
         \caption{\centering BPG-444  $28.84dB@0.2520bpp$}\label{fig:lena_bpg256}
     \end{subfigure}
     \begin{subfigure}[b]{0.22\textwidth}
        \centering
         \includegraphics[width=0.85\textwidth]{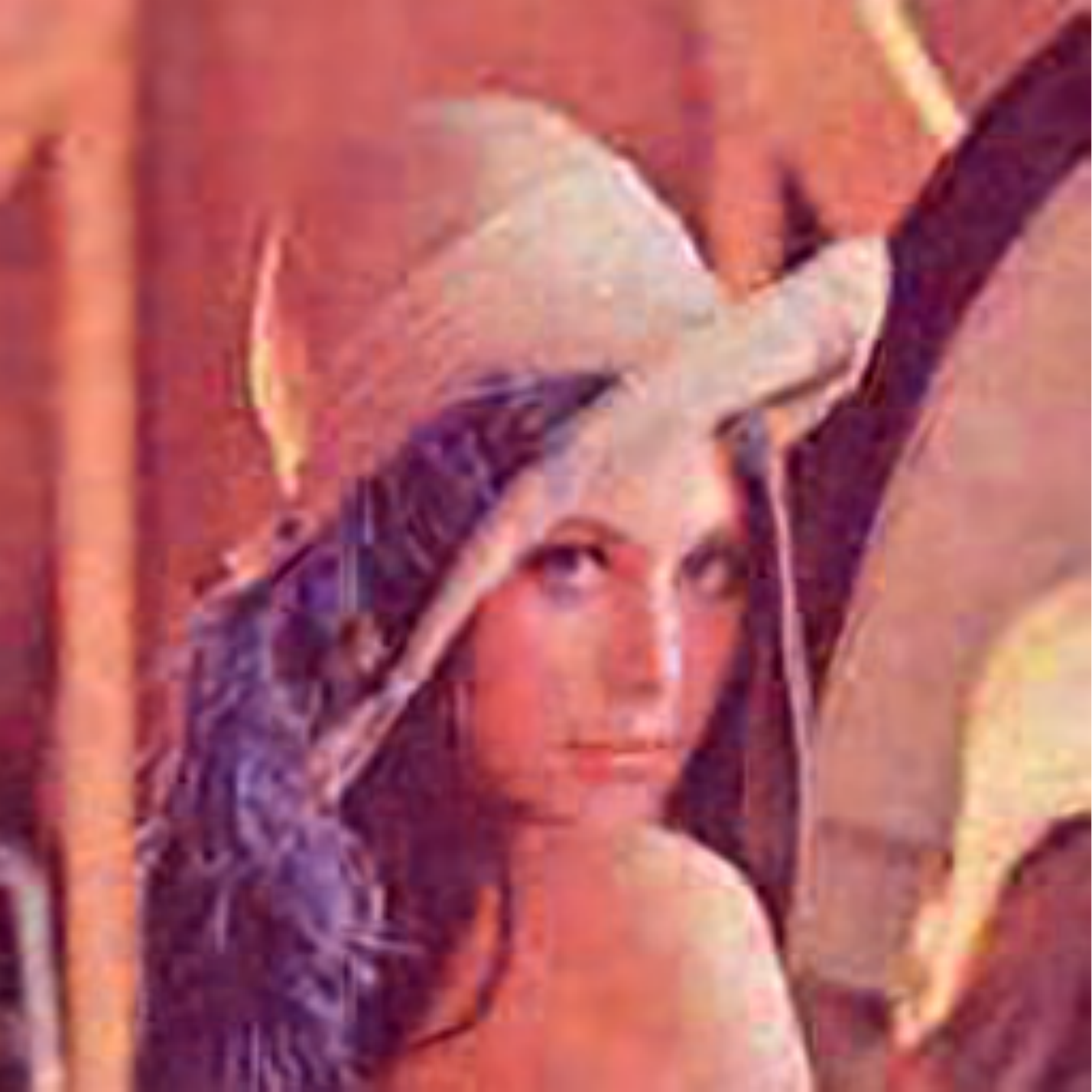}
         \caption{\centering      Our Method      $26.29dB@0.2452bpp$}\label{fig:lena_our256}
     \end{subfigure}
\caption{Evaluation results for Lena of resolution $256\times 256$.}\label{fig:lena256}
\end{figure*}

\begin{figure*}[t]
     \centering
     \begin{subfigure}[b]{0.22\textwidth}
        \centering
         \includegraphics[width=0.85\textwidth]{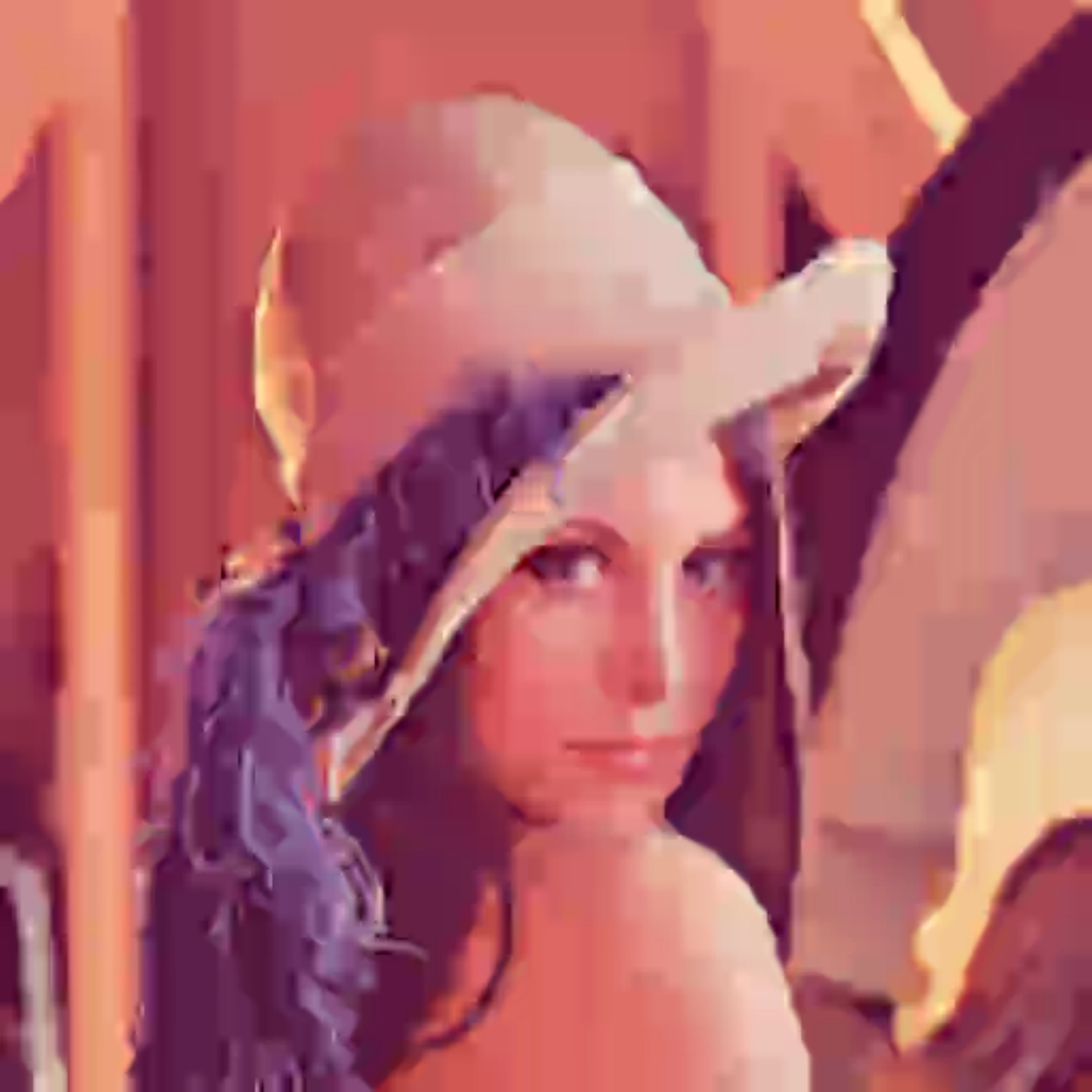}
         \caption{\centering H.264-444  $25.79dB@0.1176bpp5$}\label{fig:lena_h264512}
     \end{subfigure}
      \begin{subfigure}[b]{0.22\textwidth}
        \centering
         \includegraphics[width=0.85\textwidth]{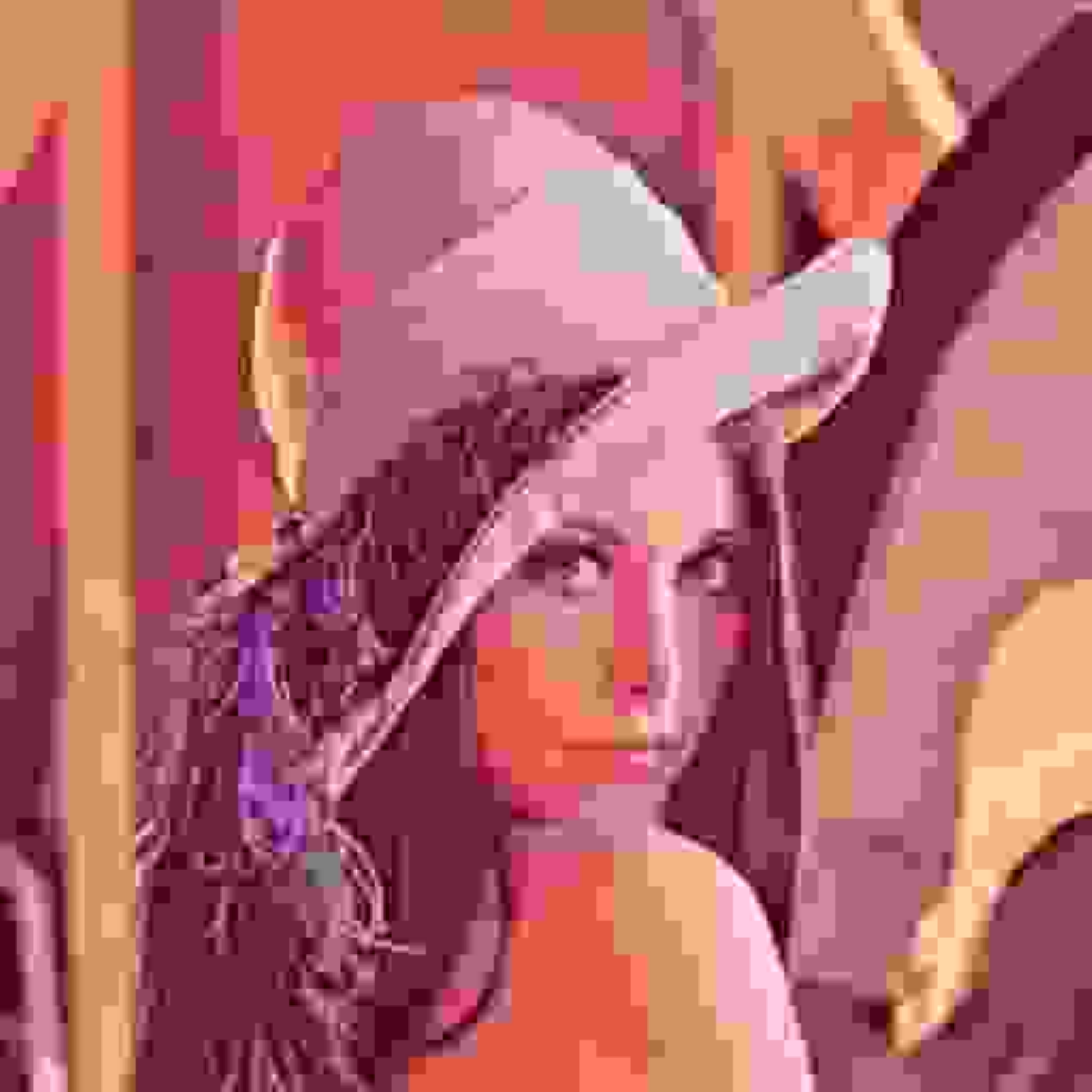}
         \caption{\centering JPEG Progressive  $21.29dB@0.1176bpp$}\label{fig:lena_jpg_prog512}
     \end{subfigure}
     \begin{subfigure}[b]{0.22\textwidth}
        \centering
         \includegraphics[width=0.85\textwidth]{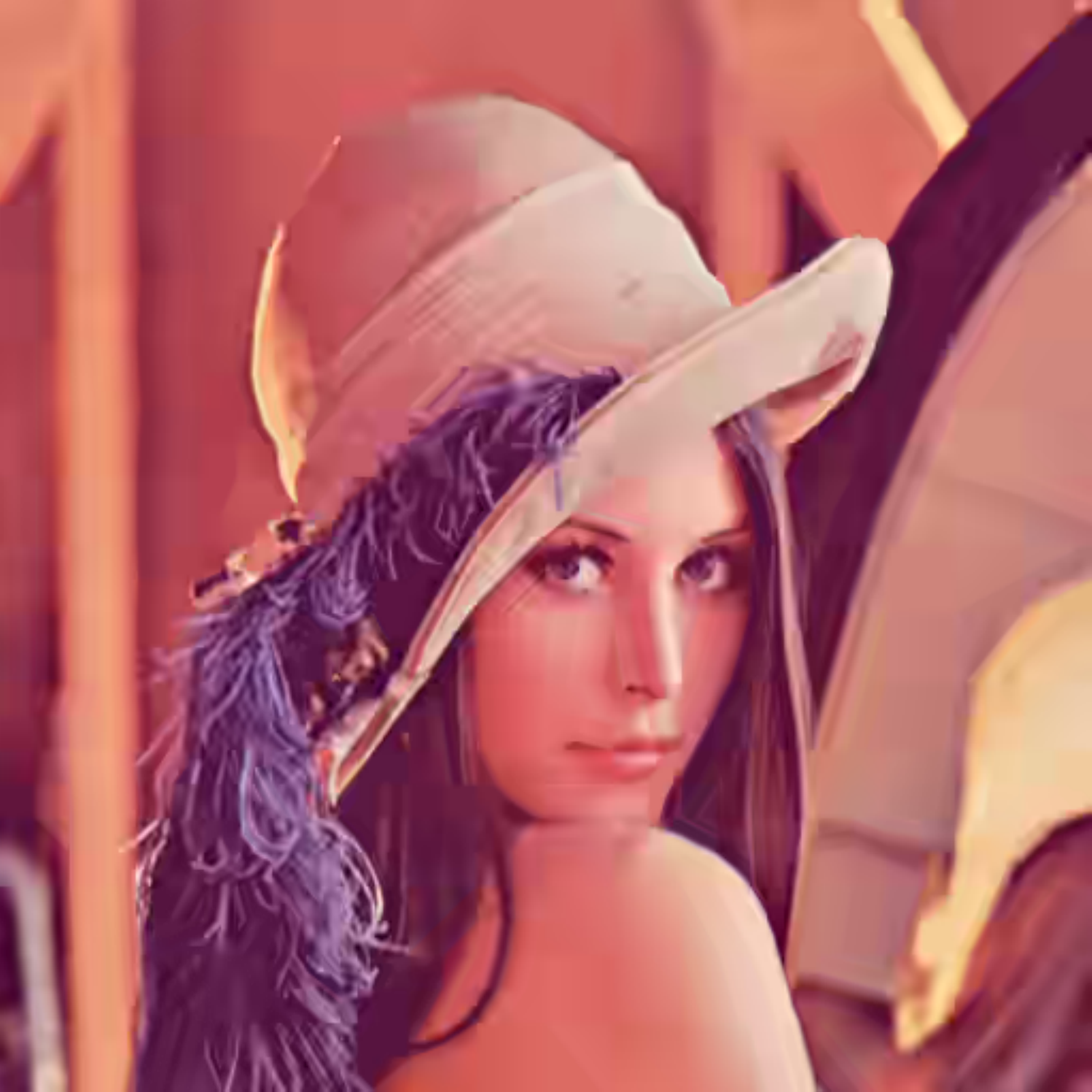}
         \caption{\centering BPG-444  $29.72dB@0.1191bpp$}\label{fig:lena_bpg512}
     \end{subfigure}
     \begin{subfigure}[b]{0.22\textwidth}
        \centering
         \includegraphics[width=0.85\textwidth]{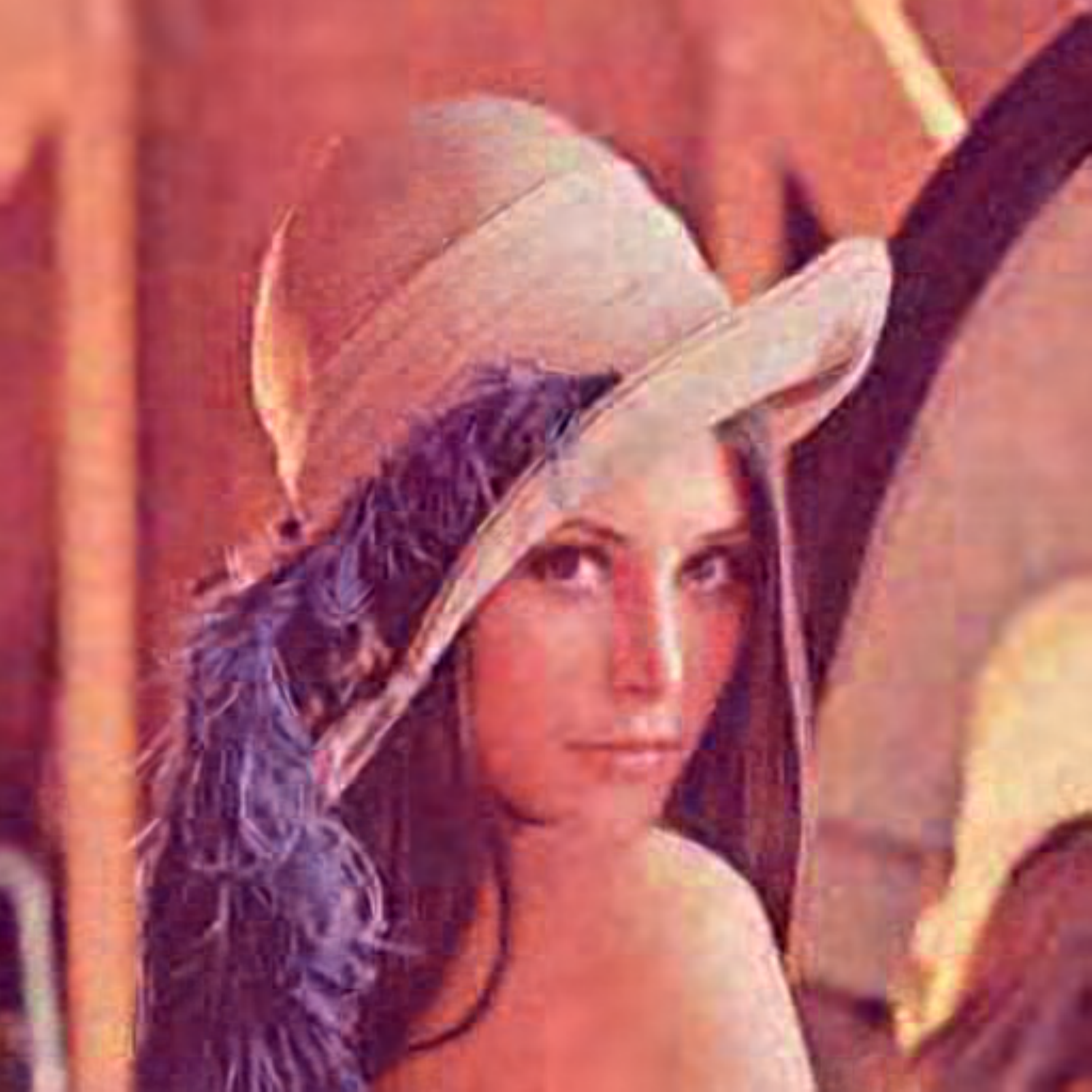}
         \caption{\centering      Our Method      $27.18dB@0.1194bpp$}\label{fig:lena_our512}
     \end{subfigure}
\caption{Evaluation results for Lena of resolution $512\times512$.}\label{fig:lena512}
\end{figure*}
    
\begin{figure*}[ht]
     \centering
     \begin{subfigure}[b]{0.48\textwidth}
        \centering
         \includegraphics[width=0.95\textwidth,height=0.6\textwidth]{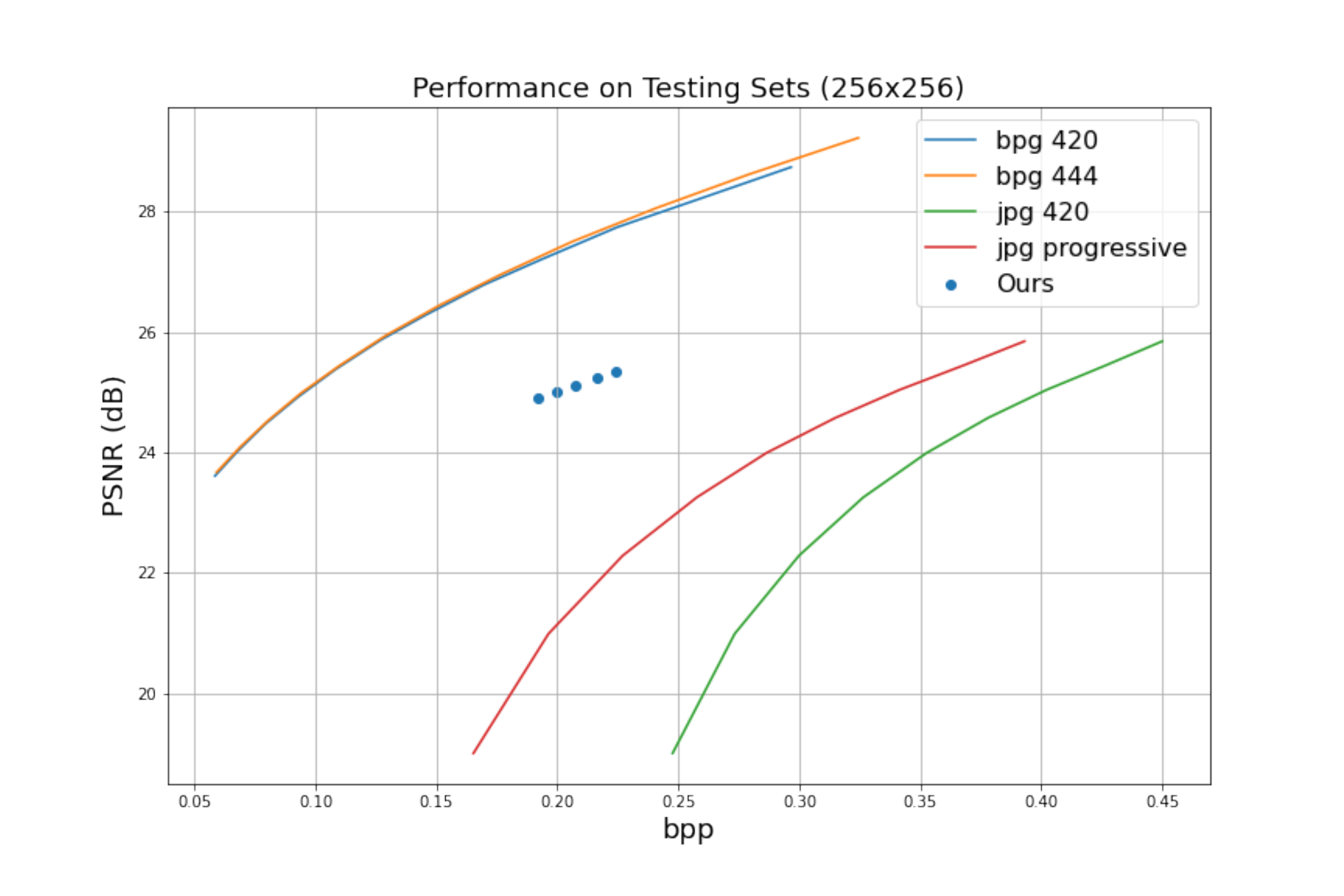}
         \caption{Scale: $256\times256$}\label{fig:clic256}
     \end{subfigure}
      \begin{subfigure}[b]{0.48\textwidth}
        \centering
         \includegraphics[width=0.95\textwidth,height=0.6\textwidth]{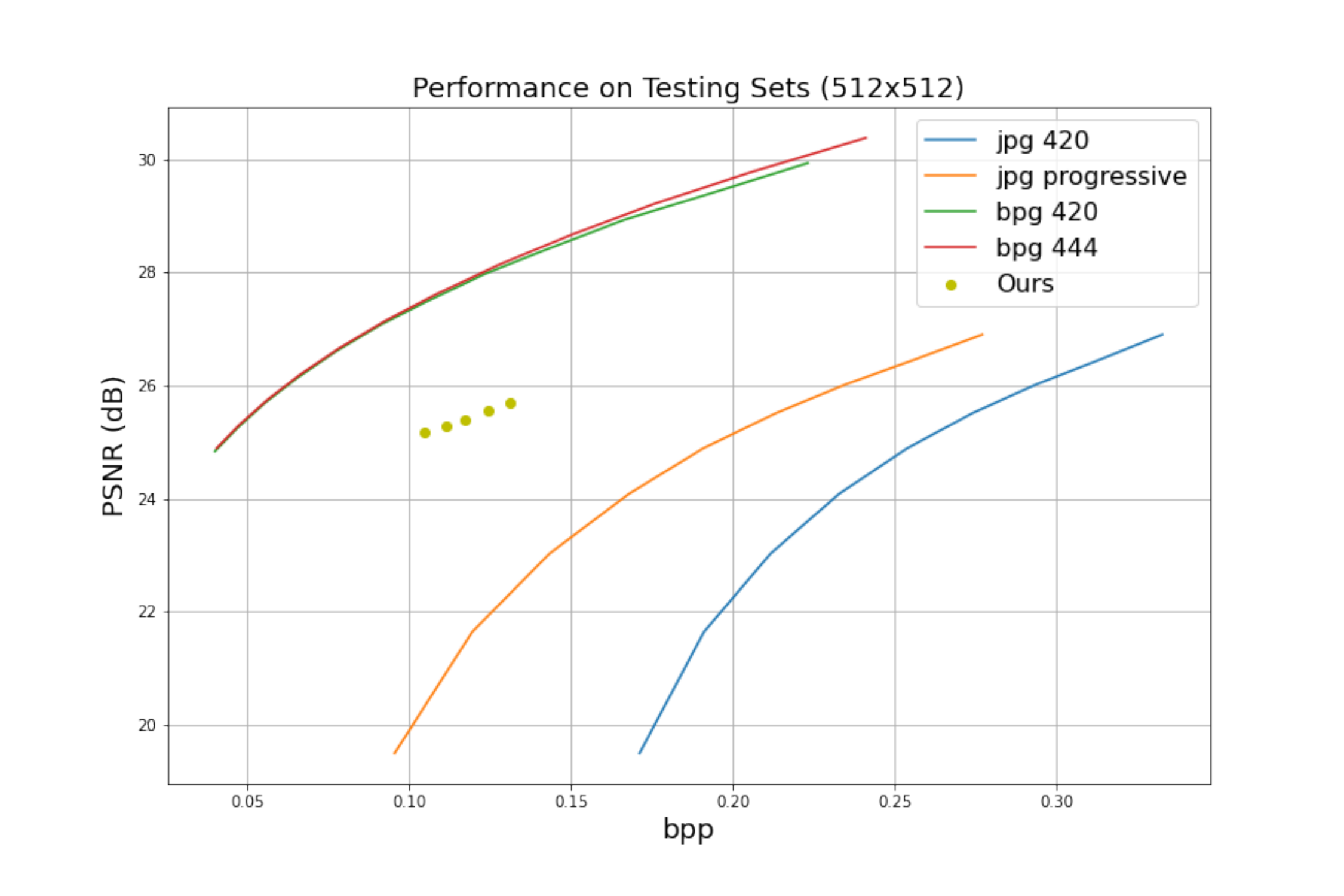}
         \caption{Scale: $512\times512$}\label{fig:clic512}
     \end{subfigure}
\caption{Evaluation results on the average of 186 test images.}\label{fig:clic_test_plot}
\end{figure*}

\begin{table}
\caption{Bit rate distribution as a function of the block size at each grid 
for Lena ($256\times256$, $26.19dB@0.2286bpp$), where QT stands for the cost 
of saving the quad-tree at each grid. Only DC in $G_2$ is coded. Other
numbers in the first column show the block size and "-" means either no bit
needed or not applicable.}\label{table:bpp_aloc}
\begin{tabular}{cccccccc}\hline
        & $G_8$     & $G_7 $    & $G_6$     & $G_5$     & $G_4$     & $G_3$ & $G_2$ \\ \hline
QT      & 4.0e-3    & 4.8e-3    & 1.3e-3    & 3.2e-4    & 7.6e-5    & 1.5e-5  &-  \\
DC  &- &- &- &- &- &- &2.4e-4\\ \hline
$C_{*,8}$     &1.5e-5     & -         & -         & -         & -         & -    &-   \\
$C_{*,7}$     & 9.1e-5    & 1.5e-5    & -         & -         & -         & -      &- \\
$C_{*,6}$      & 7.6e-4    & 1.5e-4    & 1.5e-5    & -         & -         & -     &- \\
$C_{*,5}$      & 5.1e-3    & 1.2e-3    & 1.6e-5    & 7.6e-5    & -         & -     &-  \\
$C_{*,4}$      & 0.0213    & 7.2e-3    & 1.7e-3    & 3.1e-4    & 6.1e-5    & -     &-  \\
$C_{*,3}$       & 0.0246    & 0.029     & 8.2e-3    & 2.1e-3    & 3.2e-4    & 7.6e-5 \\
$C_{*,2}$       & -         & 0.073     & 0.0322    & 8.6e-3    & 1.5e-3    & 3.2e-4 &- \\
\hline
\end{tabular}

\end{table}

\section{Experiments}\label{sec:experiment}

In the experiments, we use images from the CLIC training set \cite{x1}
and Holopix50k \cite{hua2020holopix50k} as training ones, use Lena and
CLIC test images as test ones. Except for Lena, which is of
resolution $512\times512$, we crop images of resolution $1024\times1024$
with stride 512 from the original high resolution images to serve as
training and test images. To verify the progressive characteristics, we
down-sample images of resolution $1024\times1024$ to $512\times512$ and
$256\times256$ using the Lanczos interpolation.  There are 186 test
images in total. 


We select 3400 images from the training set to train VQ codebooks. They
are trained by the faiss.KMeans method \cite{johnson2019billion}, which
provides a multi-thread implementation.  Fig. \ref{fig:rd} shows multiple codebooks $C_{n,m}$ at grids $G_n$ in terms of
the number of codewords and the number of the kept spectral components 
in the final stage in VQ training which results the parameters we used shown in TABLE.\ref{table:par256}.

Figs. \ref{fig:lena256} and \ref{fig:lena512} compare decoded images
from several standards and our MGBVQ method at the same bit rate of
resolutions $256\times256$ and $512\times512$, respectively.  Figs.
\ref{fig:clic256} and \ref{fig:clic512} compare the averaged RD
performance for 186 test images of resolutions $256\times256$ and
$512\times512$, respectively. We see from these figures that MGBVQ
achieves remarkable performance under a very simple framework without
any post-processing. More textures are retained by our framework since
the traditional transform coding scheme tends to discard high
frequencies to achieve more zeros especially when the bit rate is low.
While in our framework, high frequencies are captured by larger grid VQ
and smaller block VQ when necessary. 

The distribution of bits spent in coding at each grid is summarized in
Table \ref{table:bpp_aloc}. We see from the table that the coding of DC
and VQ indices in coarse grids account for a small percentage of the
total bit rate . The great majority of bits are spent in the finest grid
and smallest block sizes, say, the coding of indices of codebooks
$C_{8,3}$, $C_{7,2}$, $C_{7,3}$ and $C_{6,2}$. 

\section{Conclusion and Future Work}\label{sec:conclusion}

A lightweight learning-based image codec, called MGBVQ, was proposed in
this work. Its feasibility and potential has been demonstrated by
experimental results. However, there are a few details that can be
fine-tuned, including rate control and entropy coding. We will continue
to push its performance and demonstrate that the new framework will
achieve competitive performance as compared with deep-learning-based
image coding method at much lower computational complexity.

\newpage
\bibliographystyle{IEEEtran}
\bibliography{refs}

\end{document}